\documentclass[sigconf]{acmart}

\date{\today}

\title{A Generative Model of Group Conversation}

\usepackage{graphicx}
\usepackage{color}
\usepackage[section]{placeins}
\usepackage[framemethod=TikZ]{mdframed}
\usepackage{lipsum}

\usepackage{alltt}
\usepackage{tabu}

\begin{document}

\title{A Generative Model of Group Conversation} 

\author{Hannah Morrison} 
\affiliation{North Carolina State University}
\email{hmorris3@ncsu.edu}

\author{Chris Martens}
\affiliation{North Carolina State University}
\email{martens@csc.ncsu.edu}

\begin{abstract}
\noindent
Conversations with non-player characters (NPCs) in games are typically confined to dialogue between a human player and a virtual agent, where the conversation is initiated and controlled by the player. To create richer, more believable environments for players, we need conversational behavior to reflect initiative on the part of the NPCs, including conversations that include multiple NPCs who interact with one another as well as the player. 

We describe a generative computational model of group conversation between agents, an abstract simulation of discussion in a small group setting. We define conversational interactions in terms of rules for turn taking and interruption, as well as belief change, sentiment change, and emotional response, all of which are dependent on agent personality, context, and relationships. 
We evaluate our model using a parameterized expressive range analysis, observing correlations between simulation parameters and features of the resulting conversations. This analysis confirms, for example, that character personalities will predict how often they speak, and that heterogeneous groups of characters will generate more belief change.

\end{abstract}

\keywords{conversation modeling}

\maketitle

\section{Introduction}
Many digital games contain rich narrative experiences that allow for players to participate in the construction and outcomes of the story. As a part of this experience, the player typically interacts in ``first person'' by taking on a fictional role as one of the story's characters and interacting with non-player character (NPCs). From a design perspective, the construction of NPCs is a balancing act: authors generally wish to give characters {\em depth} in the form of recognizable, consistent characterization, but they also wish to allow for player agency in terms of steering their interactions with NPCs. What the player has in mind may, of course, differ from the author's intent for that character.

\begin{figure}

\begin{mdframed}[roundcorner=10pt]
{\bf Daphne:} You sure picked a spooky day to go boating, Freddy.\\
{\bf Fred:} Well it didn't start out that way. What could've happened?\\
{\bf Velma:} It's very simple. When the barometric pressure dropped, and the warm offshore air came in contact with an inland cold front, 
  we ran into some unnavigable mugilation.\\
{\bf Fred:} You're right, Velma. Whatever you said.\\
{\bf Velma:} I said, we're lost in the fog.
\end{mdframed}

\caption{An example of conversation between a group of fictional
characters from {\em Scooby Doo, Where Are You?} Season 1 Episode 3: Hassle in the Castle.}
\label{fig:scoobydoo}

\end{figure}

Such games frequently offer conversational interfaces that allow the player
to approach NPCs, strike up conversations, and make choices about what path
to take through a branching dialogue structure. However, traditionally,
underlying dialogue managers have assumed a simplified framework of a
single human player interacting with a single virtual agent in a
turn-taking manner. This assumption, while allowing for a simple authoring
model, limits the expressiveness of the system. Characters who may take
initiative to interact with the player and with one another more accurately
reflect social interactions in real life, and we hypothesize that they
serve to increase the social believability and narrative intrigue within a
game experience.

For example, Figure~\ref{fig:scoobydoo} depicts a portion of the script from an episode
of {\em Scooby Doo, Where Are You?}, a light-hearted cartoon television
series that centers around banter between a group of teenage friends, ``the
Scooby Gang,'' as they navigate spooky mysteries like haunted houses.
Conversation in existing game narratives does not yet enjoy even this kind
of banter between NPCs unless it is entirely hand-scripted, and reactive to
player actions only at pre-specified yield points. We seek to enable
interactive narrative experiences in which the player can take part in this
kind of group banter.

In service of this longer-term vision, we propose a computational model of
group conversation in the form of a generative social simulation.  The
model is {\em \bf symmetric} in the sense that it assumes the interface for
player interaction (conversational moves) is the same as the ``AI''
interface, i.e. the set of available actions for the virtual agents in the
game. In this way, the model does not predefine a locus of control, but
allows for orthogonal design decisions about where to drop in the player's
point of view within the simulation. The model is also {\em \bf generative}
in the sense that each next line of dialogue is determined by numerous
conversational variables that may be influenced by numerous prior
conversational moves. In other words, there is no pre-authored script; what
characters should say next will be decided by various features of the
conversational state. Those features include things like the current topic,
conversational obligations, character personality, affinity between
characters, and character emotions.

In this paper, we introduce our model in terms of the aspects of conversation we have chosen to represent and the rules we use to simulate them. We built our model in the Ceptre programming language, which enables us to understand causal relationships between conversational moves within generated structures, as well as to run the simulation with random variation a large number of times, giving us a set of trials from which to sample and reason about. Across these trials, we modified various input parameters of the model, such as character personality, to observe the effects that they would have on measurable qualities of conversations, such as which characters spoke more frequently.


\section{Related Work}

In order to model group discussion, we must first determine what a group is. Stodgill ~\cite{GA} defines a group as ``an open interaction system in which actions determine the structure of the system and successive interactions exert coequal effects upon the identity of the system.''  Johnson and Johnson ~\cite{JT} offer one definition of groups as ``a collection of individuals whose interactions are structured by a set of roles and norms.'' We combine these definitions for our model, operating under the definition of a group as a collection of individuals, governed by social norms, who make up a system of interaction that is affected by the actions of the group's members. 

Previous models of social interaction have used dialogue as a way to exchange information and/or services (Dixon et al, 2009; Evans, 2016). Evans~\cite{CSP} presents a model based on the Game of Giving and Asking for Reasons (GOGAR), in which agents in a discussion must justify or retract their statements if their claims are incompatible. 
The model also includes the social norms of conversational turn taking and adherence to selected topics. The model presented by Dixon et al.~\cite{PAD}, in ``introduc[ing] a framework for specifying agents who reason by planning and interact through dialogue,'' enforces the conversational norm of responding to a question or request by introducing the expectation of a response from the asker.

In the context of NPCs for games, Ryan et al.~\cite{ryan2016lightweight} created a ``a fully procedural alternative to branching dialogue'' that allows authors to tag various quips associated with characters and topics, then make use of those tags within conversational goals and plans (effectively templates). This work relates strongly to our own interest in conversational structure, since they model things like topics, moods, and conversational obligations. However, the approach differs from ours in its use of scripted conversation templates and tagging, as well as the intended application being one-on-one, player-to-character dialogue rather than group conversation. 
Short and Evans, in the {\em Versu} system~\cite{evans2014versu}, also model speech acts, turn taking, topics, and character sentiments, in the context of an interactive conversational experience with a group of fictional characters. The model we describe is quite similar, though smaller, simpler, and hopefully possible to adopt in many contexts, presented as a general framework independent from proprietary implementation.

Meanwhile, several researchers have been exploring story and dialogue modeling in linear logic. Dixon et al. represent goal-driven asking a question and receiving a response using linear logic notation, then carry out dialogue planning with linear logic proof search~\cite{PAD}. 
Bosser et al.~\cite{bosser2010linear} model potential narrative events in a story world based on {\em Madame Bovary} as linear logic propositions, where different possible proofs of a sequent represent variation in the narrative. Martens scaled this approach with the development of the linear-logic-based Ceptre language~\cite{Ceptre}, demonstrating its use for interactive story systems. These examples demonstrate the suitability of linear logic not only in modeling interactions and events, but in modeling interactions with varying outcomes.

\section{Background: Rule-Based Conversation Modeling}
The conversational rules of our model are grounded in the work of Gibson~\cite{gibson2003participation}, who states that the two certainties of conversation are that it is rule-based (in that ``who speaks and what they say are both subject to rules that ensure a basic level of order and intelligibility") and unequal (in that ``not everyone is dealt the same hand, in terms of opportunities to speak and be addressed.") In creating a conversational framework that addresses these certainties, he introduces the concept of participation shifts ({\em p-shifts}), the changing states of participation between the roles of ``speaker, target, and unaddressed recipient'' that affects each conversational participant. He categorizes these p-shifts by the ways in which a participant takes, gives, or loses one's status as a speaker. Our model reflects the rapid nature of p-shifts such that there are many ways each conversational participant may take on or cease to be in the role of speaker.

We wanted our model to be {\bf precise}, {\bf
concise}, and {\bf portable}: precision meaning that the output of the model could be understood in direct formal relationship to the code;
concision meaning that the amount of code should scale linearly with the
phenomena being modeled (and that the majority of time should be spent
implementing model-level phenomena, not underlying data structures or
bookkeeping mechanisms), and {\bf portable} meaning that the model itself
can be understood in an implementation-independent way and re-implemented
in other systems without knowing how the underlying modeling formalism
works.

For these reasons, we selected Ceptre, a linear-logic-based modeling tool. 
In Ceptre, a model is an unordered collection of declarative {\em if...
then...} rules that describe how the simulation can change components of
its state based on its current properties of its state. State components
are described by logical predicates determined by the author. The author
also provides an initial condition describing the input parameters of the
model, and the simulation runs (non-deterministically) until no more rules
can fire.

For example, in Ceptre, the following syntax describes a rule for topic
change:

\begin{verbatim}
change_topic :
    turns (N + 1) * current_topic T * relevant T T' 
    * $thinks Character (opinion T' Sentiment)
  -o
    current_topic T' * turns N.
\end{verbatim}

The \verb|*| symbol represents conjunction between predicates and \verb|-o|
is the if-then (implication) operator, in this case indicating that the
state described before it should be replaced by the state described after
it. The \verb|$| operator preserves a precondition as a result, such that
\verb|$A -o B| is equivalent to \verb|A -o A * B|.  Terms beginning with
capital letters are {\em parameters} which are implicitly quantified at the
beginning of the rule, such that we can read the preceding rules as
follows:

If there exist terms $N$, $T$, $T'$, and $C$ such that the following constraints are satisfied:
\begin{itemize}
\item There are $N+1$ turns remaining
\item The current topic is $T$
\item Character $C$ is currently speaking
\item $C$ has an opinion about topic $T'$
\item $T$ and $T'$ are related
\end{itemize}

Then the simulation state changes as follows:
\begin{itemize}
\item There are $N$ turns remaining (instead of $N+1$)
\item The current topic is $T'$ (instead of $T$)
\end{itemize}

Note that the predicates (\verb|turns|, \verb|current_topic|, and so on)
are not built-in primitives of the model but are author-declared constructs, and can be modified to represent other phenomena.

In a specific simulation state, the parameters will be instantiated with
ground (concrete) terms. For example, if there are 4 turns remaining, the
current topic is ghosts, Velma is speaking, and Velma has an opinion about
murder, which is related to ghosts, then the rule may fire with $N$ set to 3, $T$
set to ghosts, $T'$ set to murder, and $C$ set to Velma, resulting in a
state where 3 turns remain and the current topic is murder.

In any given simulation state, multiple rules may apply. For example,
someone may be able to interrupt the current speaker, or the current
speaker may be able to continue speaking. In this case, Ceptre chooses
nondeterministically between the available options, or if run in
interactive mode, the user may select which to apply. This property of the tool allows
for sampling of the generative space as well as experimentation with user
interactions with the model.

Our modeling approach does not include automated natural language generation (NLG), as we aim to separate the concerns of structural aspects of conversation from the concrete realization of lines of dialogue. Because our model produces a richly-detailed skeletal structure of conversation that we believe supports, but does not over-constrain, the associated text, we are excited to explore the possibilities for NLG that it will enable. Currently, we provide hand-interpreted natural language renderings of the conversations described, taking care only to use information available from the model when we do so.

\section{A Model of Group Conversation}

To simulate simple discussions, we define rules for conversation, emotional response,
agent personality, and belief change.  For simplicity, we designated a
leader who would begin conversation with a chosen topic, and in order to
simulate a time constraint, we imposed a 9-turn limit on conversation. A ``turn" in the context of our model provides a way for us to track and limit the amount of speech acts among participants in the interaction. Every speech act excluding initiation of conversation spends a turn.  Below we overview the aspects of conversation that we model and provide (slightly simplified) examples of the rules that implement them.

Throughout the description of the model, we will refer to examples of conversations that it generated (Figures \ref{fig:interrupt} and \ref{fig:personality}). 
The examples show the output of our system with a domain based on Scooby Doo. 
We provide a hand-rendered natural language interpretation of the transition selected by the model and a summary of the state change resulting from the transition.
The premise for these examples is that Daphne initiates a conversation about a house, where the house is related to other topics such as creaky doors and ghosts. The initial configuration establishes that Daphne and Fred hold a positive opinion of the house, while Velma holds a negative opinion of the house, and Velma and Fred have a negative opinion of the creaky doors, while Daphne has a neutral opinion of them. Additionally, we give the characters the following personality archetypes (see Section~\ref{sec:personality}): Fred is a Participant, Daphne is a People-Pleaser, and Velma is a Contrarian. 

\begin{figure}
\small

\begin{tabu} to 0.5\textwidth { | p{3.5cm} | p{2cm} | p{2cm} | }
 \hline
 \textbf{Interpreted Utterance} & \textbf{Ceptre Rule} & \textbf{State Change} \\
 \hline
\textbf{Daphne:} Let's talk about this house.   & initiate   & set topic: house; Daphne speaking \\
\hline
\textbf{Daphne:} I think it looks nice.  & finish speaking with opinion  & Daphne finished speaking \\
\hline
\textbf{Velma:} It's decrepit. I-- & begin speaking with opinion & Velma speaking \\
\hline
\textbf{Fred:} Well, I think-- & interrupt & Fred speaking; Velma is miffed \\
\hline
\textbf{Velma:} I think it looks awful.&  interrupt & Velma speaking; Fred is miffed \\
\hline

\textbf{Velma:} And people are saying it's haunted. & finish speaking with fact &  Velma finished speaking \\
\hline
\textbf{Fred:} It's very old. & begin speaking with fact & Fred speaking \\
\hline
(Daphne thinks to herself: I guess it's not as nice as I thought it was.) & positive to neutral opinion & Daphne's opinion of the house declines \\
\hline
\textbf{Fred:} It's 20 years old. & finish speaking with fact & Fred finished speaking \\
\hline
\textbf{Velma:} Fred, is the house old? It shouldn't look so worn down after such a short time. & question fact & Velma speaking \\
\hline
\textbf{Velma:} That's probably why people are saying it's haunted---but we all know it isn't. & question fact &   \\
\hline
(Fred thinks to himself: maybe Velma is right about the house.) & positive to neutral opinion & Fred's opinion of the house declines \\
\hline
\end{tabu}
\caption{An example of a model-generated sequence of dialogue moves demonstrating interruption, turn taking, questioning of statements, and sentiment change.}
\label{fig:interrupt}

\end{figure}

\subsection{Utterances}

We represent statements (declarative utterances) abstractly as facts or opinions about a topic. The term \verb|(opinion ghosts negative)|, for example, represents an utterance of a negative opinion about ghosts; or \verb|(fact ghosts make_believe)| as an utterance that ghosts are make-believe. Utterances may be held as thoughts by characters via the predicate \verb|thinks|, as in \\ 
\verb|(thinks velma (opinion ghosts negative))|, or they may be spoken aloud, which is tracked with predicates named \verb|says| and \verb|hears|.

\subsection{Conversational Structure}

Because we defined a group in part as a set of individuals governed by social norms, 
we included several rules for enforcing conversational norms. These include: only one person may speak at a time; one may only speak if their statement relates to the current topic; if one wants to change the topic, it must be relevant to the current topic.

We represent the current speaker with a predicate \verb|is_speaking C|, and we also track which characters are listening to one another with a predicate \verb|listening C C'| (by default we assume all characters are listening to one another, though this is something we would like the flexibility to change later).

Turn-taking conversation is mainly modeled with the {\em begin speaking}, {\em finish speaking}, and {\em change topic} rules. The rule for changing topics was given above; here is a rule for {\em begin speaking}:

\begin{verbatim}
begin_speaking :
    $current_topic T * $listening C C' * 
    finished_speaking C' * thinks C Statement * 
    on_topic Statement T * turns (N+1)
  -o
    is_speaking C * hears C' C Statement * turns N.
\end{verbatim}

This rule requires that, if the character \verb|C| is listening to \verb|C'|, then \verb|C'| must be finished speaking, and \verb|C| must have something to say that pertains to the current topic. For example, any opinion of the form 
\verb|(opinion T S)| is on-topic for the topic \verb|T|.\footnote{This relationship between statements and topics is defined in a separate part of the model as a permanent fact that does not change during simulation.}

In group discussion, however, sometimes these norms are violated. In order
to add some realism to our model, we defined a rule for interruption, which
represents someone interrupting and continuing to talk over the current
speaker. This disrupts the normal flow of conversation, will affect some
group member's emotional states, and, depending on what is said by the
interrupter, may result in the belief change of one or more group members. 
The following rule permits interruption in our model.

\begin{verbatim}
interrupt :
   turns (N+1) * $is C Type * 
   interruptive Type * is_speaking C' * 
   $listening C C'
 -o
   interrupts C C' * is_speaking C * 
   feels C' miffed * turns N. 
\end{verbatim}

This rule requires that \verb|C|'s personality archetype \verb|Type| is classified as interruptive and that some other character \verb|C'|, to whom \verb|C| is listening, is speaking. As a consequence, we record that \verb|C| has interrupted \verb|C'|, change the current speaker to \verb|C|, add an emotional state (\verb|miffed|) to \verb|C'|, and spend a turn.

See Figure~\ref{fig:interrupt} for an example of a generated conversation in which both turn-taking conversation and interruption have taken place.

\subsection{Emotional Response}

What is heard during a conversation might affect another agent's emotional
state, which in turn might affect their speech. While we have not
explicitly modeled speech affected by emotion, these effects could perhaps be rendered in
natural language based on the existence of emotional predicates when a conversational rule fires. Future work may explore how agents are affected by different emotional
deliveries of dialogue.

We represent emotion with a predicate \verb|feels C F|, where \verb|C| is a character and \verb|F| is an emotion.
Rules that change emotions include: becoming happy if one agrees with someone;
becoming sad if one disagrees with someone; becoming encouraged to speak if
one is involved in the conversation multiple times; becoming annoyed if one
is interrupted, and angry if one is interrupted multiple times. To give a concrete example, the rule for becoming angry after enough interruption is:

\begin{verbatim}
upset_from_interruption :
   feels C miffed * feels C miffed
  -o
   feels C angry.
\end{verbatim}

This rule includes the \verb|feels C miffed| predicate twice as a precondition, which in linear logic is distinct from the predicate appearing just once, representing a larger quantity of that emotional resource.



\subsection{Agent Personality}
\label{sec:personality}

\begin{figure}
\small

\begin{tabu} to 0.5\textwidth { | p{3.5cm} | p{2cm} | p{2cm} | }
 \hline
 \textbf{Interpreted Utterance} & \textbf{Ceptre Rule} & \textbf{State Change} \\
 \hline
\textbf{Daphne:} Well, here we are in the lakeside cabin!  & initiate   & set topic: house  \\
\hline
\textbf{Daphne:} Isn't it beautiful?  & finish speaking with opinion &  Daphne finished speaking \\
\hline
\textbf{Fred:} It's kind of old and musty,  & begin speaking with fact & Fred is speaking\\
\hline
\textbf{Fred:} but it looks nice, considering.  & finish speaking with opinion & Fred finished speaking \\
\hline
\textbf{Daphne:} Oh, do you really think so, Fred? & question opinion & Daphne questioned Fred \\
\hline

\textbf{Velma:} People are saying it's haunted.   & begin speaking with fact &  Velma is speaking \\
\hline
\textbf{Velma:} Jinkies!
& finish speaking with opinion 
& Daphne hears a negative opinion about the house \\
\hline
\textbf{Daphne:} Well, I guess it {\em is} a little spooky.& agree to please & Velma hears a negative opinion of the house \\
\hline
\textbf{Velma:} Daphne, it's frankly hideous, too. & cause debate & Daphne hears a negative opinion of the house \\
\hline
\textbf{Daphne:} Uh, yeah. I guess so.  
& agree to please & Velma hears a negative opinion of the house \\
\hline
\end{tabu}

\caption{An example of a model-generated sequence of dialogue moves demonstrating turn taking, agreeing to please, and causing debate.}

\label{fig:personality}

\end{figure}

Not all conversants will be emotionally affected by speech in identical ways, or participate in identical manners. In order to add realism to the model, we introduced four personality archetypes for the agents that determined which rules of the model would apply to them, in turn affecting their participation in the conversation.
%
{\bf Participant} agents represent the prototypical conversational participant, someone who will speak actively and frequently about what they believe, who will occasionally talk over someone in order to get their point across. 
{\bf People-pleaser} agents represent a participant who is concerned with what the other participants think of them. They will be happy if someone agrees with them, or sad if someone disagrees with them, and may echo what others say despite disagreeing in order to maintain the approval of the group. They will not talk over others.
{\bf Contrarian} agents, like their participant-type counterparts, will participate actively and may speak over others, however contrarians may introduce an opposing opinion for the sake of keeping the conversation going.  
Finally, {\bf reticent} agents are reluctant to speak, and must be directly engaged before they will contribute to the conversation. Like their people-pleaser counterparts, they do not talk over others.

These archetypes affect conversational rules by being required as additional constraints on certain conversational moves.
Two examples of rules influenced by character personality are ``agreeing to please'' and ``causing debate." The ``agree to please'' rule allows a people-pleaser to state an opinion that aligns with a previously spoken one, even if they do not actually hold that opinion:

\begin{verbatim}
agree_to_please:
    turns (s N) * hears C C' (opinion T S) *
    $current_topic T * $is C people_pleaser * 
    $listening C C' * $listening C' C
  -o
    says C (opinion T S) * hears C' C (opinion T S) 
    * turns N.
\end{verbatim}

An example of this rule being used in a generated conversation, as well as the \verb|cause_debate| rule, can be found in Figure~\ref{fig:personality}.

\subsection{Sentiment Change}

In order to simulate realistic, dynamic discussion, we defined rules for
agent sentiment change change. An agent's opinion on a topic may be negative,
neutral, or positive. For simplicity's sake, an agent's opinion will go
from positive to neutral, or negative to neutral, if they hear an opposing
opinion on the topic. An agent's opinion will go from neutral to negative,
or neutral to positive, if the agent hears an opposing opinion on the topic
and has a favorable opinion of the person saying it. For example,
the rule for changing opinion from negative to neutral is: 

\begin{verbatim}
negative_to_neutral_opinion :
   $current_topic T * thinks C (opinion T negative) * 
   hears C C'(opinion T positive)
-o
   thinks C (opinion T neutral).
\end{verbatim}

This rule states that when the topic is current, when \verb|C| holds a negative
opinion on the topic and hears \verb|C'| say something positive about the topic,
then \verb|C|'s opinion of the topic changes to neutral. An example of this rule in the context of a generated conversation can be found in Figure~\ref{fig:interrupt}.

\subsection{Visualizing Causality}

\begin{figure}
\centering
\includegraphics[scale=.25]{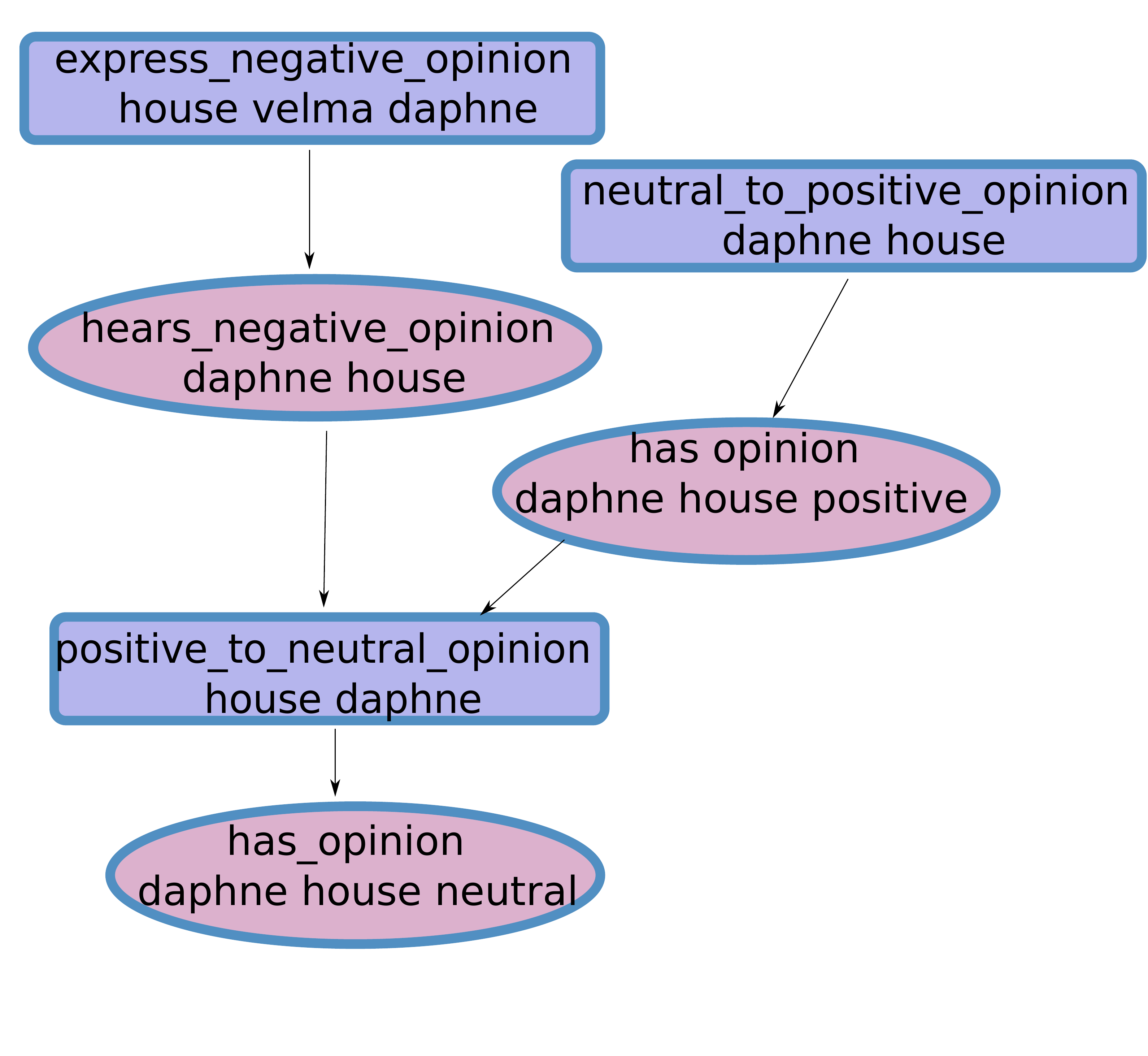}

\caption{A simplified depiction of belief change, where rectangular nodes represent 
a rule that fired, and oval nodes represent the resources that are generated and
consumed by rules.}

\label{fig:causal}

\end{figure}

By representing our conversation rules in terms of their dependencies and outputs,
we obtain conversational structures that not only give rise to a {\em sequence} of dialogue moves, 
but also reveal the causal interconnections. For example, as seen in Figure~\ref{fig:causal}, we can see that both Velma expressing a negative opinion of the house within Daphne's earshot, and Daphne's previous sentiment shift about the house from neutral to positive, led to a sentiment shift back to neutral. The two preceding events are not causally-linked, however---we can reason about these graphs as a partial ordering on simulation events, permitting possibilities for reordering, just as similar structures do for narrative plots~\cite{riedl2010narrative}.

\section{Analysis}

\begin{figure}
\includegraphics[scale=.8]{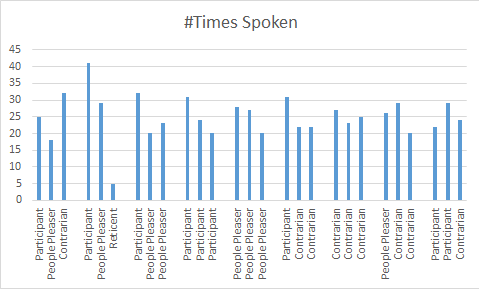}
\caption{The total number of times each participant spoke for 9 combinations of participant groups.}
\label{fig:timesspoke}
\end{figure}

\begin{figure}
\includegraphics[scale=.8]{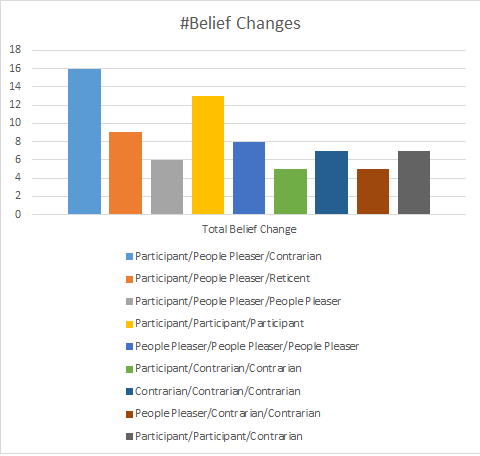}
\caption{The total number of belief changes that occurred across all participants for 9 combinations of participant groups.}
\label{fig:beliefchanges}
\end{figure}

Since we have not yet built an automated NLG pipeline for our model, we have not evaluated it
with human interactors. However, we have attempted to understand the expressive range~\cite{smith2010analyzing} of our system by varying the initial conditions, running many iterations of the nondeterministic program, and observing features of the output to look for correlations between input parameters and output features. This type of qualitative understanding may be considered a ``parameterized expressive range'' evaluation.

For this work, the parameters we focused on were the group composition, namely the personality archetypes of all characters in the conversation. We wanted to know whether the {\em prescription} of these archetypes matched a {\em functional} understanding of them---e.g., does a character described as ``reticent'' actually speak less? We were also interested in whether certain group compositions
were more likely to lead to a high frequency of belief change.

Therefore, we ran simulations using varied group compositions, running each 15 times. 
We automated trace processing, adding to a tally of belief change each time a belief-change
rule fired, and adding to the tally of times spoken every time a speaking
rule for a specific character fired. We found that varying agent personality archetypes 
resulted in different levels of belief change and differing distributions of
conversational participation.

Figure~\ref{fig:timesspoke} depicts the variation in conversational participation due to differing combinations of agent archetypes. Conversation is most one-sided in the Participant/People-Pleaser/Reticent combination, and, interestingly, most balanced when all participants are Contrarians. We suspect that contrary group composition would be less equitable if we built in the idea that characters' emotions may affect their willingness to speak. 

Figure~\ref{fig:beliefchanges} depicts the variation in belief change due to differing combinations of agent archetypes. The highest amount of belief change occurs in the Participant/People-Pleaser/Contrarian combination, which we interpret as the rules specific to the People-Pleaser and Contrarian archetypes leading to characters voicing their opinions more frequently, in turn leading to belief change. In other words, our model is one where more heterogeneous groups of characters will lead to a more dynamic narrative, a property that seems to align with presiding theories of drama (see, e.g., the concept of a {\em foil}~\cite{auger2010anthem}).


\section{Conclusion}

We have presented a precise, concise, and portable model of group 
interaction implemented using Ceptre. The complete executable
ruleset is available on GitHub.\footnote{\url{https://github.com/chrisamaphone/group-conversation/}}
Running simulations with this model demonstrates that it is capable of producing varied traces 
that resemble social conversation, and that the structure of conversation produced reflects the personality traits and initial conditions for conversation in a predictable, yet varied, way.


In future work, we intend to create applications of the model, such as interactive experiences,
by adopting a natural language pipeline that makes use of the skeletons produced by the model.
We also intend to evaluate the believability and coherence of the generated conversations through human subjects. 

We hypothesize that to create more comprehensible conversations, we will need to implement some goal-directed behavior. It would be interesting to explore the use of backward-chaining search (as in planning) over characters' conversational goals to supplement our forward-chaining approach; as in narrative at large, we suspect that human readers will look for intentionality in characters' conversational behavior~\cite{riedl2004intent}.

With support for goal-driven behavior, the model could include rules that deal with more complex
relationships between group members (e.g. two members that will continually
support each other's opinions because of a friendship) and add nuance to the agents' process of
belief change, as well as the effects of agent emotion on speech and agent reception of emotional speech.

\bibliographystyle{ACM-Reference-Format}
\bibliography{main}













\end{document}